\RequirePackage{etoolbox}
\patchcmd{\bibliographystyle}{#1}{unsrtnat}{}{}
\PassOptionsToPackage{numbers,sort&compress}{natbib}
\documentclass{ecai} 

\usepackage{latexsym}
\usepackage{amssymb}
\usepackage{amsmath}
\usepackage{amsthm}
\usepackage{enumitem}

\usepackage[utf8]{inputenc} 
\usepackage[T1]{fontenc}    
\usepackage{hyperref}       
\usepackage{url}            
\usepackage{booktabs}
\usepackage{amsfonts}       
\usepackage{nicefrac}       
\usepackage{microtype}      
\usepackage[dvipsnames]{xcolor}         
\usepackage{colortbl}
\usepackage{nicematrix}
\usepackage{tcolorbox}
\usepackage{graphicx}
\usepackage[font=footnotesize]{caption}
\usepackage{subcaption}
\usepackage{wrapfig}
\usepackage{float}
\usepackage{commath}

\usepackage{selectp}

\definecolor{gray1}{gray}{0.85}
\definecolor{gray2}{gray}{0.89}
\definecolor{gray3}{gray}{0.93}

\newcommand{\inlineBox}[2]{\tcbox[size=fbox, colback=#1,colframe=#1,on line,boxsep=0pt]{#2}}
\newcommand{\inlineBoxPad}[2]{\tcbox[size=fbox, colback=#1,colframe=#1,on line,boxsep=1pt]{#2}}

\newcommand{\BibTeX}{B\kern-.05em{\sc i\kern-.025em b}\kern-.08em\TeX}

\graphicspath{{images/}}

\newcommand{\better}[1]{\textcolor{Green}{#1}}
\newcommand{\worse}[1]{\textcolor{Red}{#1}}

\newcommand{\colorU}[1]{\textcolor[HTML]{810f7c}{#1}}
\newcommand{\colorG}[1]{\textcolor[HTML]{007D81}{#1}}

\newcommand{\lossU}{\colorU{\mathcal{L}_{\mathit{u}}}}
\newcommand{\lossG}{\colorG{\mathcal{L}_{\mathit{g}}}}

\newcommand{\cart}{\texttt{CartPole-v1}}
\newcommand{\acro}{\texttt{Acrobot-v1}}
\newcommand{\pendu}{\texttt{InvertedPendulum-v4}}
\newcommand{\pusher}{\texttt{Pusher-v4}}
\newcommand{\hopper}{\texttt{Hopper-v4}}
\newcommand{\walker}{\texttt{Walker2d-v4}}
\newcommand{\cheetah}{\texttt{HalfCheetah-v4}}
\newcommand{\ant}{\texttt{Ant-v4}}

\begin{document}


\begin{frontmatter}


\paperid{0340} 


\title{Local Pairwise Distance Matching for Backpropagation-Free Reinforcement Learning}




\author{Daniel Tanneberg}
\address{\includegraphics[scale=0.08]{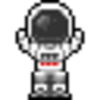} Honda Research Institute EU
}


\begin{abstract} 
Training neural networks with reinforcement learning (RL) typically relies on backpropagation (BP), necessitating storage of activations from the forward pass for subsequent backward updates. 
Furthermore, backpropagating error signals through multiple layers often leads to vanishing or exploding gradients, which can degrade learning performance and stability.
We propose a novel approach that trains each layer of the neural network using local signals during the forward pass in RL settings.
Our approach introduces local, layer-wise losses leveraging the principle of matching pairwise distances from multi-dimensional scaling, enhanced with optional reward-driven guidance. 
This method allows each hidden layer to be trained using local signals computed during forward propagation, thus eliminating the need for backward passes and storing intermediate activations. 
Our experiments, conducted with policy gradient methods across common RL benchmarks, demonstrate that this backpropagation-free method achieves competitive performance compared to their classical BP-based counterpart. 
Additionally, the proposed method enhances stability and consistency within and across runs, and improves performance especially in challenging environments. 
\end{abstract}
\end{frontmatter}

\section{Introduction}
\label{sec:intro}
Deep reinforcement learning (DRL) has achieved remarkable advances in recent years, trained primarily by stochastic gradient descent and leveraging the power of large models and extensive data~\cite{mnihHumanlevelControlDeep2015,silverMasteringGameGo2016,silverMasteringGameGo2017,vinyalsGrandmasterLevelStarCraft2019,wangDeepReinforcementLearning2024a}.
While methods like tree-search have supplemented these techniques~\cite{silverMasteringGameGo2016}, the backbone of training these large neural networks remains end-to-end learning using backpropagation (BP)~\cite{rumelhartLearningRepresentationsBackpropagating1986}.

Despite its pivotal role in the success of deep (reinforcement) learning~\cite{lecunDeepLearning2015,wangDeepReinforcementLearning2024a}, BP has known limitations.
It is not considered a biologically plausible learning paradigm for the cortex, although some efforts have been made to model how real neurons might implement it ~\cite{scellierEquilibriumPropagationBridging2017,lillicrapBackpropagationBrain2020,hintonForwardForwardAlgorithmPreliminary2022a}. 
Additionally, backpropagation requires to store all activations, entails delayed updates, demands computationally expensive full backward passes, and requires full knowledge of the model. 
These challenges have motivated the exploration of alternative training methods that do not rely on BP, an area of research that dates back to earlier studies~\cite{bartoGradientFollowingBackpropagation1987,bengioGreedyLayerWiseTraining2007,wierstraNaturalEvolutionStrategies2014} and has gained renewed interest in recent years~\cite{salimansEvolutionStrategiesScalable2017,baydinGradientsBackpropagation2022,hintonForwardForwardAlgorithmPreliminary2022a,ororbiaPredictiveForwardForwardAlgorithm2023,guan2024temporal}.

Several approaches have been developed to enable backpropagation-free training of neural networks. 
These include (or combine) alternative gradient calculation methods during the forward pass~\cite{bartoGradientFollowingBackpropagation1987,jaderbergDecoupledNeuralInterfaces2017a,baydinGradientsBackpropagation2022,renScalingForwardGradient2023}, layer-wise (pre-)training ~\cite{hintonFastLearningAlgorithm2006,bengioGreedyLayerWiseTraining2007,wersingLearningOptimizedFeatures2003a,kulkarniLayerwiseTrainingDeep2017}, using local information~\cite{hintonFastLearningAlgorithm2006,noklandTrainingNeuralNetworks2019}, backward passes of targets ~\cite{bengioHowAutoEncodersCould2014,leeDifferenceTargetPropagation2015,ororbiaBiologicallyMotivatedAlgorithms2019,meulemansTheoreticalFrameworkTarget2020}, black-box optimization~\cite{wierstraNaturalEvolutionStrategies2014,salimansEvolutionStrategiesScalable2017,tannebergEvolutionaryTrainingAbstraction2020}, and dedicated frameworks or models~\cite{kostasAsynchronousCoagentNetworks2020,hosseiniHierarchicalPredictiveCoding2020,ororbiaNeuralCodingFramework2022,millidgePredictiveCodingFuture2022,ororbiaBackpropFreeReinforcementLearning2022,graetzInfomorphicNetworksLocally2023,guan2024temporal}.

In this paper, we propose a novel backpropagation-free approach that combines local loss functions with layer-wise training in a reinforcement learning (RL) setting. 
RL imposes an additional challenge as there is no fixed dataset.
Instead, the training data is generated dynamically by the current model parameters, and all layers of the network contribute to this process, necessitating continuous updates. 
Therefore, we leverage the concept of matching pairwise distances from multi-dimensional scaling (MDS)~\cite{torgersonMultidimensionalScalingTheory1952,steyversMultidimensionalScaling2002,saeedSurveyMultidimensionalScaling2018}, a well-established non-linear dimensionality reduction technique~\cite{leeNonlinearDimensionalityReduction2007}. 
Layers are trained with a loss that aims to preserve the pairwise distances in the input data at its output, relying on information available at the training layer during the forward pass, eliminating the need for backward passes.

A similar concept of matching pairwise similarities was explored in~\cite{noklandTrainingNeuralNetworks2019}, though it was applied in supervised learning settings and utilized additional neural structures. 
Additionally, predicting random distances has been used for unsupervised representation learning~\cite{wangUnsupervisedRepresentationLearning2020} for anomaly detection and clustering. 
Most layer-wise training approaches update each layer until convergence or add new layers while freezing previous ones~\cite{hintonFastLearningAlgorithm2006,bengioGreedyLayerWiseTraining2007,wersingLearningOptimizedFeatures2003a,kulkarniLayerwiseTrainingDeep2017,hintonForwardForwardAlgorithmPreliminary2022a,ororbiaPredictiveForwardForwardAlgorithm2023}. 
While suitable for (un)supervised learning with a static dataset, this is challenging in RL settings, where the training data is generated online and depends on the full current network's interaction with the environment.

A notable exception to the predominant supervised learning applications is found in~\cite{ororbiaBackpropFreeReinforcementLearning2022}, which presents a method that fits within frameworks utilizing dedicated neural models like predictive coding~\cite{hosseiniHierarchicalPredictiveCoding2020,millidgePredictiveCodingFuture2022,ororbiaNeuralCodingFramework2022}. 
These methods primarily aim for biological plausibility and are not compatible with classical neural networks. 
In contrast, our proposed approach can be easily integrated into classical neural networks and is compatible with established RL algorithms.

In summary, we introduce a method that defines a local loss for each hidden layer based on matching pairwise distances. 
This method requires no backward pass or memorization of activations and is compatible with classical networks and any RL algorithm. 
We propose two variations of this pairwise distance-based local loss, with the additional possibility of integrating domain or task-specific distance measurements. 
Our experiments evaluated this approach with established policy gradient algorithms -- REINFORCE~\cite{williamsSimpleStatisticalGradientfollowing1992}, REINFORCE with learned baseline~\cite{mei2022role}, and PPO~\cite{schulmanProximalPolicyOptimization2017} -- across a set of common RL benchmark environments. 
The results demonstrate that our backpropagation-free approach can compete with classical backpropagation-based training, and can enhance performance and stability in many cases.

\begin{figure*}[t!]
	\centering
	\includegraphics[width=0.9\linewidth]{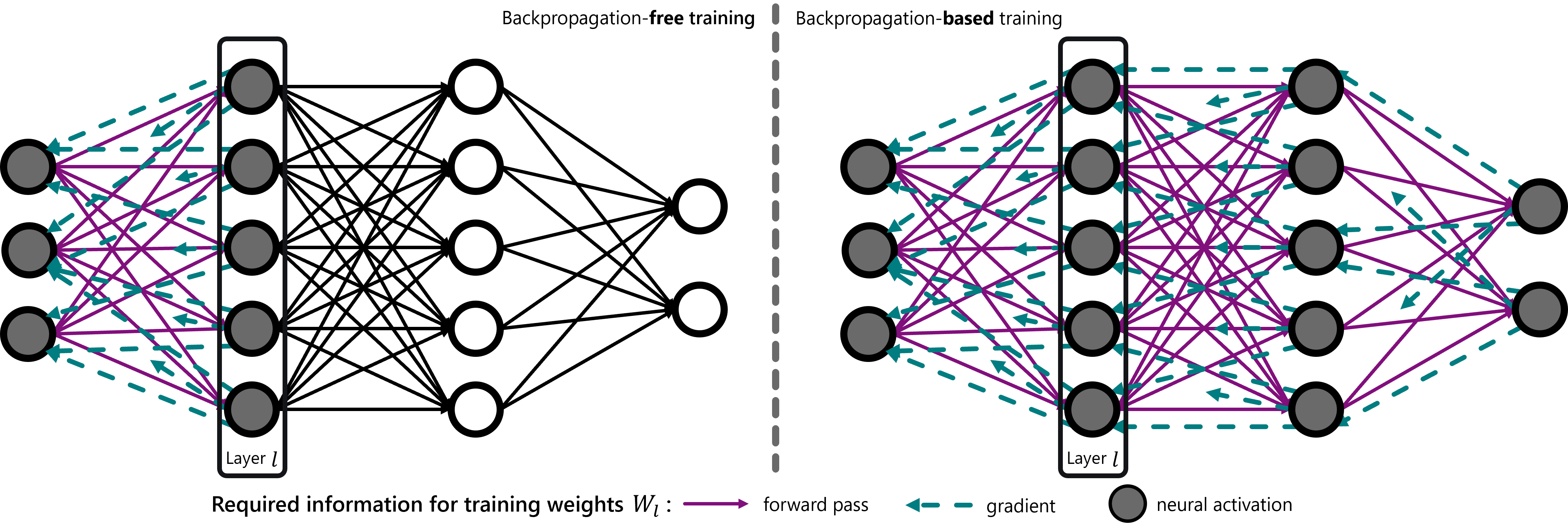}
	\caption{
		Overview of the required information in classical backpropagation-based training (\textbf{right}) and the proposed backpropagation-free (\textbf{left}) approach using local signals when training the weights $W_l$ of layer $l$. 
		Note that for clarity not all arrows are depicted. 
	}
	\label{fig:networks}
	\vspace{5pt}
\end{figure*}

\section{Preliminaries}
Here, we provide a brief introduction into important concepts used in the paper.
For a more detailed description we refer to existing literature, for example~\cite{koberReinforcementLearningRobotics2013,wangDeepReinforcementLearning2024a,saeedSurveyMultidimensionalScaling2018}.

\subsection{Markov Decision Process and Reinforcement Learning}
In reinforcement learning an agent interacts with the environment and receives a reward depending on its performance on a defined task.
The aim of the agent is to learn a reward-maximizing behavior, known as policy.
The problem is typically formalized as a Markov Decision Process (MDP) and represented as a tuple $(\mathcal{S},\mathcal{A},\mathcal{R},\gamma,\mathcal{P})$, where $\mathcal{S}$ represents the set of environment states, $\mathcal{A}$ is the set of agent actions, $\mathcal{R}: \mathcal{S} \times \mathcal{A} \rightarrow \mathbb{R}$ is the reward function, $\gamma$ is the discount factor, and $\mathcal{P}: \mathcal{S} \times \mathcal{A} \times \mathcal{S} \rightarrow \mathbb{R}$ is the state transition probability distribution function.
To learn a policy $\pi: \mathcal{S} \rightarrow \mathcal{A}$, that describes how to behave in a state, the agent aims to maximize the expected cumulative discounted reward (or return $G_t$), i.e., 
$\pi^* = \arg\!\max_\pi \mathbb{E}_{a_t\sim\pi(\cdot|s_t),s_{t+1}\sim\mathcal{P}(\cdot|s_t,a_t)} \big[ G_t \big]$, with $G_t = \sum_{t=0}^{\infty} \gamma^t \mathcal{R}(s_t, a_t)$.

\subsection{Policy Gradient Methods}
Policy gradient methods optimize parameterized policies with respect to the return using gradient descent, i.e., $\theta = \theta + \alpha \nabla_\theta J$ with learning rate $\alpha$.
The stochastic policy $\pi$ is parameterized by $\theta$, typically a neural network called \textit{actor}, and $J(\theta) = \mathbb{E}_\pi [G_t]$ is the objective that maximizes the return.
Utilizing Markov Chain Monte-Carlo (MCMC) to approximate the expectation with samples and the policy gradient theorem, the episodic policy gradient algorithm REINFORCE~\cite{williamsSimpleStatisticalGradientfollowing1992} is given by $\nabla_\theta J = \mathbb{E}\big[ \sum_{t=1}^T (G_t - b) \nabla_\theta \log \pi_\theta (a_t | s_t) \big]$, with an arbitrarily chosen baseline $b$ to reduce variance and stabilize training~\cite{williamsSimpleStatisticalGradientfollowing1992,petersReinforcementLearningMotor2008}.
The baseline can be state dependent and learned, \textit{state value baseline}, which reduces the aggressiveness of the update~\cite{mei2022role}.
Often layers are shared for learning the policy and the baseline.

\subsection{Actor-Critic Methods}
Actor-critic methods~\cite{grondmanSurveyActorCriticReinforcement2012} use a second parametrization $\phi$, again typically a neural network called \textit{critic}, to learn the baseline.
Often \textit{actor} and \textit{critic} network share the hidden layers and the two learning objectives are weighted accordingly, similar as for learned baselines.
The \textit{actor} chooses an action $a_t$ in a given state $s_t$, whereas the \textit{critic} informs the \textit{actor} how good the action was.
Using the advantage function $A(s_t, a_t) = Q(s_t, a_t) - V(s_t)$ for the \textit{critic}, which can be estimated, for example, with temporal difference error as $A(s_t, a_t) = \mathcal{R}(s_t, a_t) + \gamma V(s_{t+1}) - V(s_t)$, the Advantage Actor Critic (A2C)~\cite{mnihAsynchronousMethodsDeep2016} algorithm is realized as $\nabla_\theta J = \mathbb{E}\big[ \sum_{t=1}^T A(s_t, a_t) \nabla_\theta \log \pi_\theta (a_t | s_t) \big]$.
Proximal Policy Optimization (PPO)~\cite{schulmanProximalPolicyOptimization2017} adds the idea of trust-region updates by restricting the policy update to stay \textit{close} to the old policy and using a replay buffer with the concept of importance sampling for multiple updates, formally as $J^{\mathit{CLIP}}(\theta) = \mathbb{E}_t\big[ \min(r_t(\theta) A(s_t, a_t), \textrm{clip}(r_t(\theta), 1 - \epsilon, 1 + \epsilon) A(s_t, a_t) ) \big]$ with $r_t(\theta) = \frac{\pi_{\theta}(a_t | s_t)}{\pi_{\theta_{\mathit{old}}}(a_t | s_t)}$ the ratio indicating the change with respect to the old policy.

\subsection{Multi-dimensional Scaling}
Multi-dimensional scaling (MDS) is a technique for non-linear dimensionality reduction~\cite{torgersonMultidimensionalScalingTheory1952,leeNonlinearDimensionalityReduction2007,saeedSurveyMultidimensionalScaling2018}, typically used to map high-dimensional data onto a low-dimensional representation for analyzing and visualization.
There are different realizations of MDS, using different distance or similarity measures or cost functions to optimize~\cite{steyversMultidimensionalScaling2002,sunExtendingMetricMultidimensional2011,saeedSurveyMultidimensionalScaling2018}.
The core idea, however, remains the same.
Given a distance matrix with pairwise distances between the data points and a chosen dimension $N$ of the low-dimensional representation, MDS tries to place each data point in the low-dimensional space such that the distances are preserved, i.e., maintaining the structure of the data.

\begin{figure*}[t!]
	\includegraphics[width=\linewidth]{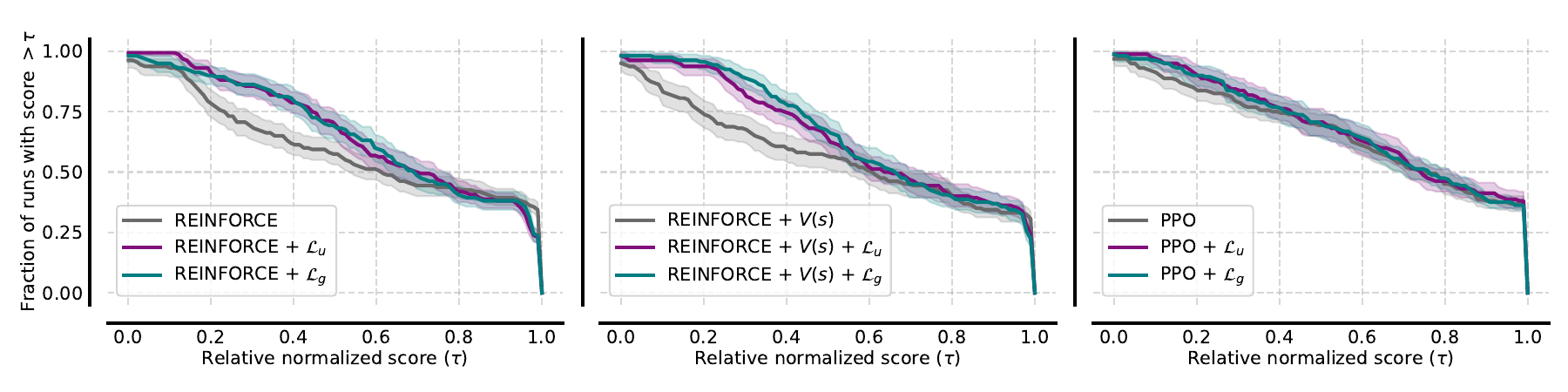}
	\caption{
		Performance profiles~\cite{agarwalDeepReinforcementLearning2021} of the proposed backpropagation-free method, $+ \lossU$ and $+ \lossG$ respectively, and their backpropagation-based baseline.
		Solid lines show the score distributions and shaded areas show pointwise $95\%$ percentile stratified bootstrap CIs.
	}
	\label{fig:profiles}
	\vspace{5pt}
\end{figure*}

\section{Local Pairwise Distance Matching} 
\label{sec:proposed_method}
Here, we first provide a high-level overview of the proposed method, before explaining the local losses in more detail.
The proposed method is based on the observation that in neural networks the hidden layers typically learn higher-dimensional feature transformations, i.e, transforming the input data into higher-dimensional spaces and by this the network forms a hierarchy of features, that (ideally) make the decisions at the last layer \textit{easier}.
On the other hand, dimensionality reduction techniques try to map higher-dimensional data onto lower-dimensional representations.
Reversing this mapping process, a local loss function for each layer can be formulated, that learns to map input data into higher-dimensional feature spaces, which enables the output layer to efficiently learn a policy.
Importantly, these local loss functions for the hidden layers do not require backpropagation and only use local signals -- see Figure~\ref{fig:networks} for a comparison of the required information in the proposed backpropagation-free training in contrast to classical backpropagation-based training.

Often, layers in neural networks \textit{increase} the dimensionality, so we want to \textit{reverse} the process of (non-linear) dimensionality reduction techniques, and learn a mapping from lower to higher dimensions -- layers can also decrease the dimensionality, but here the same idea holds.
Thus, our approach is that each hidden layer is learning a mapping from one feature space into another feature space with the constraint that pairwise distances should be preserved -- the \textit{structure} of the data should be preserved after the feature transformation.
Using only the pairwise distances between input and layer output, the learned feature transformations are unsupervised and task-agnostic, hence, enable a straightforward approach that may be beneficial for transfer, multi-task, meta, and multi-agent learning.
Additionally, the pairwise distance loss can be enhanced with reward information, such that the learned feature transformations can incorporate information about the performance and are \textit{guided} towards more \textit{useful} transformations.
\vspace{-5pt}
\begin{tcolorbox}[size=small, colback=gray3, colframe=gray3]
While the hidden layers of the network are trained with the proposed layer-wise pairwise distance loss, the output layer can be trained with any suitable reinforcement learning algorithm.
\end{tcolorbox}
\vspace{-5pt}
In the experiments we combined our approach with REINFORCE~\cite{williamsSimpleStatisticalGradientfollowing1992} without and with state value baseline ($+ V(s)$), and PPO~\cite{schulmanProximalPolicyOptimization2017}.
Note, in contrast to unsupervised pre-training~\cite{botteghiUnsupervisedRepresentationLearning2022,xiePretrainingDeepReinforcement2022,zhaoRepresentationLearningContinuous2022}, our approach works \textit{online} and always trains the whole network in a single training loop.
Next we describe the proposed local loss in detail.

\subsection{Unsupervised Pairwise Distance Loss}
\label{sec:pair_loss}
For hidden layers, we define the layers' local loss with respect to the input, noted as matrix $X = [x_0, .., x_N]^\mathsf{T}$ of $N$ stacked $\mathit{in}$-dimensional input vectors $x_n \in \mathbb{R}^{\mathit{in}}$.
A pairwise distance matrix $D_X = \{ {d_{i,j}~|~i, j \in N} \}$ is constructed by calculating the distances between all input vectors $x_i$, i.e., $d_{i,j} = \| x_i - x_j \|_1$.
We used the $\ell_1$-norm as distance measurement as it is better suited for higher dimensions~\cite{aggarwalSurprisingBehaviorDistance2001} and worked best in preliminary tests, but other distance measurements are possible and may be adapted depending on the domain or model. 
The distance matrix $D_X$ is normalized as $\bar{D}_{X} = D_X / \max{D_X}$, i.e., distances are normalized between $[0,1]$ to reflect relative distances.

The $\mathit{out}$-dimensional output of a layer $l$ for a given input $h_i^{l-1}$ is given by $y_i^{l} = \mathit{act} ( W_l^\mathsf{T} h_i^{l-1} )$, with $W_l$ the weights of the layer and a non-linear activation function $\mathit{act}$ -- here we used $\mathit{tanh}$ as activation function if not stated differently.
Using the layers outputs $y_i^{l}$, the output distance matrix $\bar{D}_{Y}$ is created similar to $\bar{D}_{X}$.
Importantly, the gradient between layers is stopped, i.e., using \texttt{pytorch}-like notation: $y_i^{l} = \mathit{act} ( W_l^\mathsf{T} \texttt{detach(}h_i^{l-1}\texttt{)} )$.

The hidden layers loss function is defined as minimizing the \textit{distance} between distance matrices, i.e., the learned transformation of layer $l$ should reflect the pairwise distances in the data.
In other words, the structure of data should be preserved by matching the pairwise distances as in MDS.
To learn the weights $W_l$, we optimize the 

\begin{tcolorbox}[size=small, colback=gray3,colframe=gray3]
	\textit{unsupervised} loss $\lossU$:
	\vspace{-8pt}
	\begin{equation}
		\label{eq:loss}
		\min_{W_l} \lossU(\bar{D}_{X}, \bar{D}_{Y}) = { \|\bar{D}_{X} / \mathit{out} - \bar{D}_{Y} / \mathit{in} \|_F} \quad, 
	\end{equation}	
\end{tcolorbox}

where $\|\cdot\|_F$ is the Frobenius norm, and both matrices are scaled by their respective data dimensionality to counter the curse of dimensionality of distances in high-dimensional spaces, i.e., guiding the learned distances to separate in high-dimensional spaces and not collapse to a relative small cluster.
The proposed local loss Eq.~\ref{eq:loss} does not require backpropagation or other forms of backward passes, only forward passes are used during inference and training.

The optimization is done with stochastic gradient descent similar like \textit{standard} training of neural networks, as
\begin{equation}
	\label{eq:gradDescent}
	W_l = W_l + \alpha \nabla_{W_l}\lossU(\bar{D}_{X}, \bar{D}_{Y})  \quad,
\end{equation}
with learning rate $\alpha$ for the Adam optimizer~\cite{kingmaAdamMethodStochastic2015}.

This training procedure is completely unsupervised and task-independent, making it suitable for any task and setup, and the learned hidden layers, i.e., the learned feature transformations, can be directly transferred to other tasks (in the same domain) -- making it interesting, for example, for multi-agent setups and meta learning.

\subsection{Guided Pairwise Distance Loss}
Additional information like prior knowledge or rewards may be included when constructing $\bar{D}_{X}$ to guide the feature learning.
The target pairwise distances can be modified with such additional information by scaling the distances accordingly.
One possible scaling is to add a data point dependent (learned) value to the respective rows and columns of $D_X$.
Given the (learned) values $v_i$\footnote{In the experiments we used the normalized return (REINFORCE), the critic's state value (PPO), and the learned baseline ($+ V(s)$) respectively.}, we first shift and normalize them to be in the range $[0,1]$ with $v_i = v_i - \min_i{v_i}$ followed by $v_i = v_i / \max_i{v_i}$.
Next, we reverse the values to put more focus on the worse performing states by increasing their distances, and divide them by two such that the total added values for each entry in the distance matrix is in $[0,1]$, i.e., $v_i = (1 - v_i) / 2$.
With ${D_X}$ being the original input distance matrix and $v$ the vector of transformed state values, the modified distance matrix ${\hat{D}_X}$ is created as \inlineBox{gray1}{${\hat{D}_X} = D_X + v + v^\mathsf{T}$} and is normalized as before as $\hat{D}_{X} = \hat{D}_X / \max{\hat{D}_X}$.
This modified distance matrix is used to create a loss similar to the unsupervised loss $\lossU$ in Eq.~\ref{eq:loss} but with \textit{performance} information as the
\begin{tcolorbox}[size=small, colback=gray3,colframe=gray3]
	\textit{guided} loss $\lossG$:
	\vspace{-8pt}
	\begin{equation}
		\label{eq:loss_g}
		\min_{W_l} \lossG(\inlineBox{gray1}{$\hat{D}_{X}$}, \bar{D}_{Y}) = { \|\inlineBox{gray1}{$\hat{D}_{X}$} / \mathit{out} - \bar{D}_{Y} / \mathit{in} \|_F} \quad,
	\end{equation}	
\end{tcolorbox}
and is optimized as before with gradient descent as in Eq.~\ref{eq:gradDescent}.
Similar as different distance measurements can be used to create the distance matrices, different modifications based on performance-based measurements can be incorporated in different ways to create different guidance using Eq.~\ref{eq:loss_g}.

\section{Experiments}
\label{sec:experiments}
We evaluated the proposed method in $8$ RL benchmarks using \texttt{gymnasium}~\cite{towers2024gymnasium} and \texttt{mujoco}~\cite{todorovMuJoCoPhysicsEngine2012}.
The used environments span from rather simpler and discrete action spaces, e.g., \cart, up to more complex and continuous action spaces, e.g., \cheetah.
To show the compatibility with different RL algorithms, we evaluated our approach with and against the following algorithms: REINFORCE~\cite{williamsSimpleStatisticalGradientfollowing1992} without and with state value baseline ($+ V(s)$), and PPO~\cite{schulmanProximalPolicyOptimization2017}.
For each environment, we compared the three baseline algorithms using backpropagation-based training (updating all layers with the RL algorithm) against the combination of the RL algorithm (updating only the output layer) with our proposed backpropagation-free local losses $+ \lossU$ and $+ \lossG$ respectively.

\begin{wraptable}{r}{4cm}
	\vspace{-15pt}
	\caption{
		Parameters.
	}
	\label{table:parameters}
	\centering\footnotesize
	\begin{tabular}{l c}
		\toprule
		parameter & value \\
		
		\midrule
		
		network size  & $[128,256]$ \\   
		non-linearity & $\mathit{tanh}$ \\ 
		$\gamma$      & $0.99$          \\ 
		$\sigma^2$    & $0.1$  \\ 
		$\alpha$      & $3e^{-4}$ \\ 
		$g_{\mathit{clip}}$ & $1$ \\ 
		$w_{\mathit{val}}$ & $0.5$ \\ 
		\midrule
		\multicolumn{2}{c}{PPO specific} \\
		\midrule
		$\mathit{replayBuffer}$ & $2000$ \\ 
		$\mathit{batchSize}$ & $128$ \\
		$\mathit{epochs}$ & $10$ \\
		$\epsilon$ & $0.2$ \\
		$w_{\mathit{KL}}$ & $0.01$ \\
		$\epsilon_{\mathit{value}}$ & $0.5$ \\
		\bottomrule
	\end{tabular}
	\vspace{-5pt}
\end{wraptable}

\begin{figure*}[t!]
	\centering
	\begin{subfigure}[b]{0.93\textwidth}
		\centering
		\includegraphics[width=\textwidth]{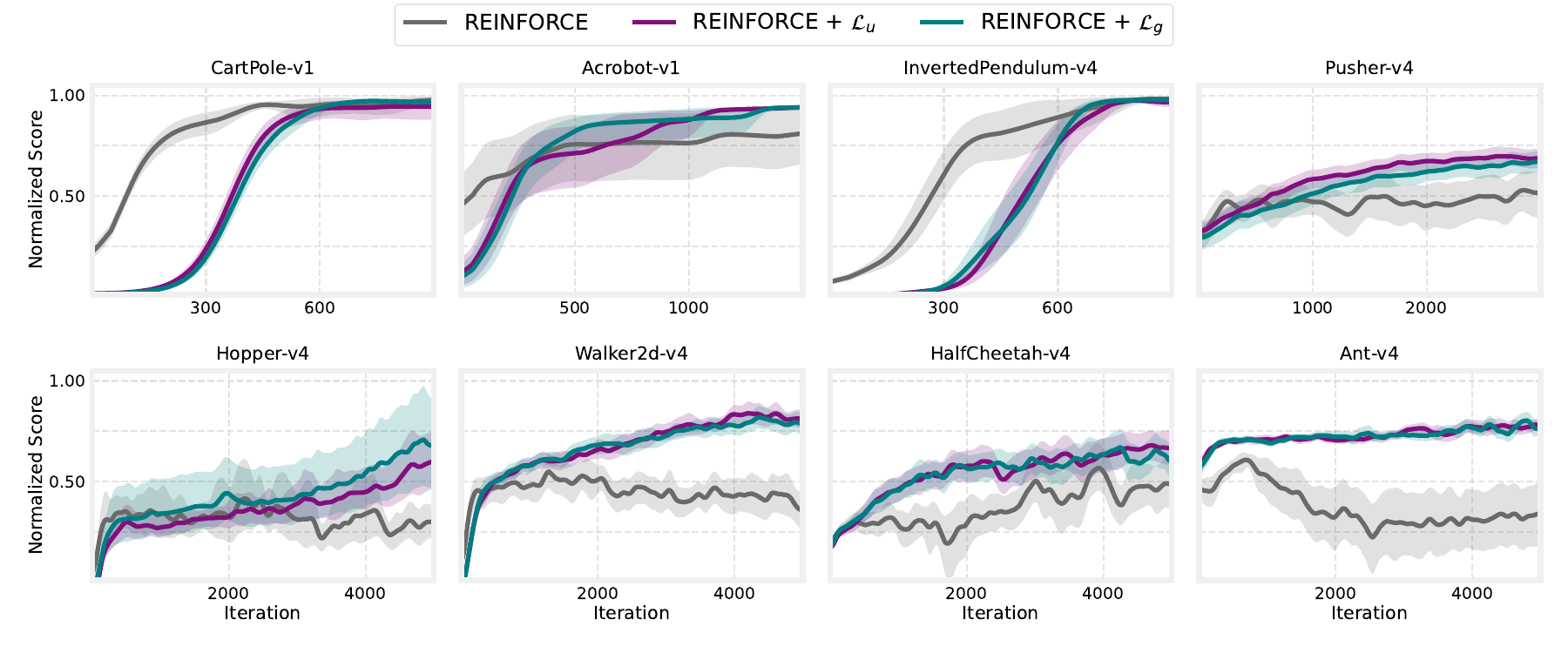}
		\vspace*{-20pt}
		\caption{Comparing against and with REINFORCE.}
		\vspace*{15pt}
		\label{fig:eval_reinf}
	\end{subfigure}
	
	\begin{subfigure}[b]{0.93\textwidth}
		\centering
		\includegraphics[width=\textwidth]{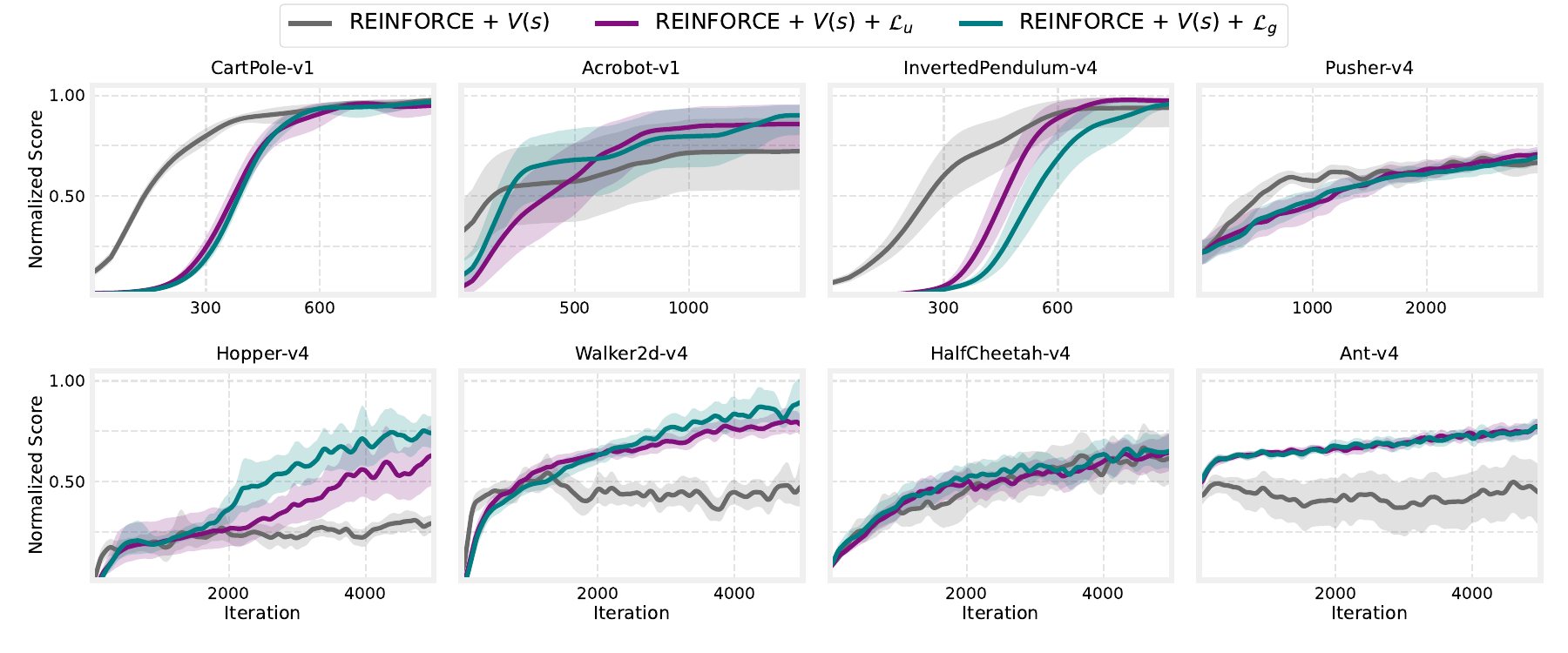}
		\vspace*{-20pt}
		\caption{Comparing against and with REINFORCE $+ V(s)$.}
		\vspace*{15pt}
		\label{fig:eval_a2c}
	\end{subfigure}
	\begin{subfigure}[b]{0.93\textwidth}
		\centering
		\includegraphics[width=\textwidth]{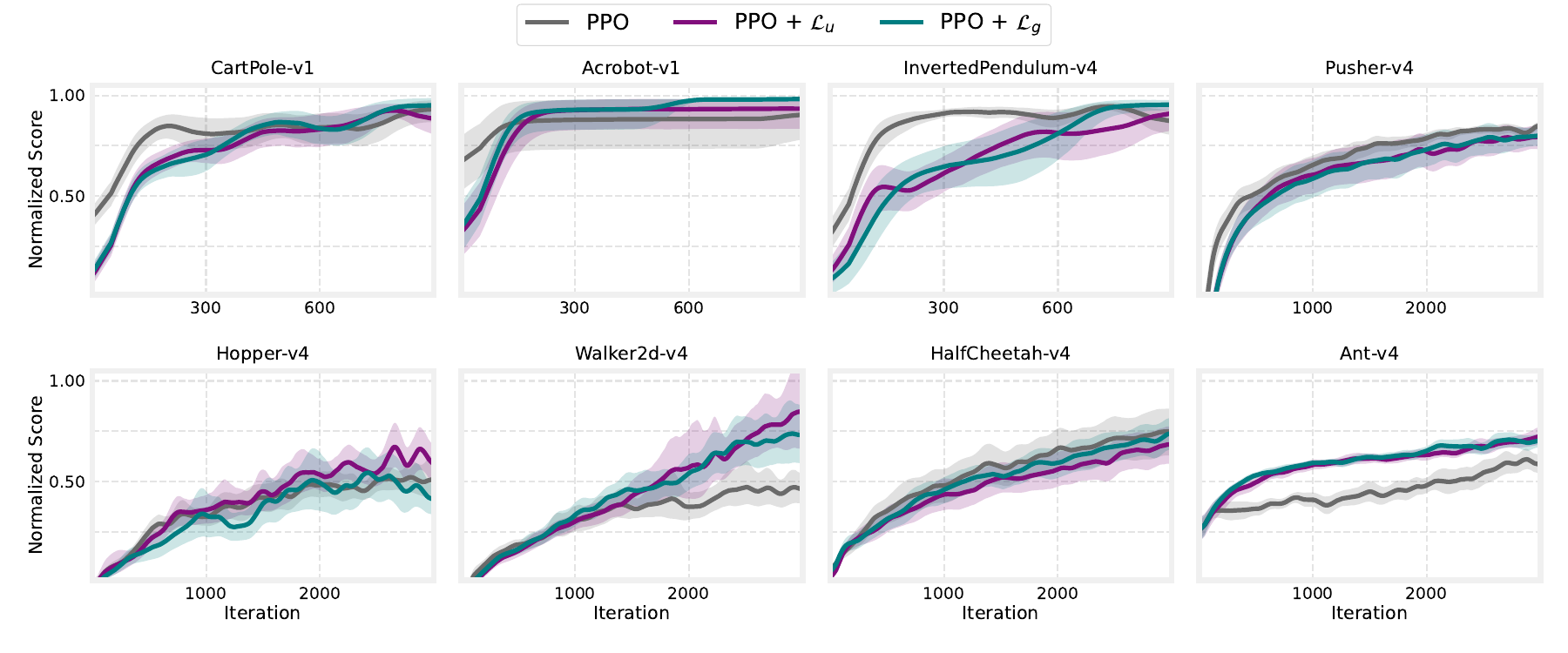}
		\vspace*{-20pt}
		\caption{Comparing against and with PPO.}
		\vspace*{15pt}
		\label{fig:eval_ppo}
	\end{subfigure}
	
	\caption{
		Comparing (a) REINFORCE, (b) REINFORCE $+ V(s)$, and (c) PPO with backpropagation against the proposed backpropagation-free method with the two loss variations, $+ \lossU$ and $+ \lossG$ respectively. 
		Plots show the mean (solid line) and $95\%$ CI (shaded area) of the normalized score over $20$ runs for each method and smoothed over $100$ iterations.
	}
	\label{fig:eval}
\end{figure*}

\subsection{Setup \& Parameters}
\label{sec:setup_parameters}
For all experiments we use the settings and hyperparameters listed in Table~\ref{table:parameters}, with the following exceptions.
Learning rate $\alpha$ was set to $5e^{-4}$ for \cart, \acro, and \pendu;
to $1e^{-4}$ for \pusher, and \cheetah~(PPO);
and to $5e^{-5}$ for \ant~(PPO).
Variance for continuous actions $\sigma^2$ set to $0.05$ for \pusher, \ant, and \cheetah~(PPO).
All other parameters were kept fixed, with $g_{\mathit{clip}}$ for gradient clipping and discount factor $\gamma$. 

For REINFORCE $+ V(s)$ and PPO, we used a shared network for the \textit{actor} and \textit{critic} with $w_{\mathit{val}}$ weighting of the value loss for the backpropagation baseline, the proposed method trains all layers independently and does not need this weighting.
The additional PPO hyperparameters are $\epsilon$ for clipping, $w_{\mathit{KL}}$ the KL penalty weighting, and $\epsilon_{\mathit{value}}$ for clipping the \textit{critic} objective.
REINFORCE and REINFORCE $+ V(s)$ use episode-based learning, i.e., one episode = one iteration, while PPO uses a replay buffer for updates (one iteration = one PPO update using \#$epochs$).
Here we use a basic PPO implementation without optimizations~\cite{shengyi2022the37implementation} and improvements like GAE~\cite{schulman2015high}, i.e., the PPO performance does not reflect its best known performance.
Importantly here, however, we are interested in the relative performances and both the backpropagation baseline as well as our approaches use this same implementation.
All parameters were set by empirical pre-evaluations and were not optimized for individual environments or algorithms as the scope of this paper is not creating new state of the art results, but to compare the relative performance between backpropagation-based and -free training.
Hence, we used the same settings for all setups (environment + algorithm), with the aforementioned variations.

\begin{table*}[h!]
	\caption{Comparing the proposed backpropagation-free methods ($+ \lossU$ and $+ \lossG$) with and against the backpropagation-based baselines of REINFORCE (REI.), REINFORCE $+ V(s)$ (REI. $+ V(s)$), and PPO.
		\inlineBoxPad{gray1}{\textit{Max score}} indicates the maximal reached cumulative reward averaged over $100$ iterations, \inlineBoxPad{gray2}{\textit{Rel. spread}} measures the relative spread around this maximal score as the $95\%$ CI in relation to the maximal score (in \%), and \inlineBoxPad{gray3}{\textit{Rel. iter.}} shows the iteration in which the maximal score was reached (for the baselines) and when this maximal score was matched by the proposed methods (if it was not reached, it is just their respective best iteration).
		The relative changes are colored \better{\textit{green}} if the proposed method was at least as good as the baseline and \worse{\textit{red}} if it was more than \textit{1.1x} worse.
	}
	\label{table:results}
	\vspace{0pt}
	\centering\footnotesize
		\begin{NiceTabular}{p[l]{1.45cm} *{4}{p{0.8cm}p{0.94cm}p{0.8cm}}}[cell-space-top-limit = 3pt,cell-space-bottom-limit = 0pt]
				\CodeBefore
				\rectanglecolor{gray1}{2-2}{11-2} 
				\rectanglecolor{gray1}{13-2}{21-2}
				\rectanglecolor{gray1}{2-5}{11-5}
				\rectanglecolor{gray1}{13-5}{21-5}
				\rectanglecolor{gray1}{2-8}{11-8}
				\rectanglecolor{gray1}{13-8}{21-8}
				\rectanglecolor{gray1}{2-11}{11-11}
				\rectanglecolor{gray1}{13-11}{21-11}
				\rectanglecolor{gray2}{2-3}{11-3} 
				\rectanglecolor{gray2}{13-3}{21-3}
				\rectanglecolor{gray2}{2-6}{11-6}
				\rectanglecolor{gray2}{13-6}{21-6}
				\rectanglecolor{gray2}{2-9}{11-9}
				\rectanglecolor{gray2}{13-9}{21-9}
				\rectanglecolor{gray2}{2-12}{11-12}
				\rectanglecolor{gray2}{13-12}{21-12}
				\rectanglecolor{gray3}{2-4}{11-4} 
				\rectanglecolor{gray3}{13-4}{21-4}
				\rectanglecolor{gray3}{2-7}{11-7}
				\rectanglecolor{gray3}{13-7}{21-7}
				\rectanglecolor{gray3}{2-10}{11-10}
				\rectanglecolor{gray3}{13-10}{21-10}
				\rectanglecolor{gray3}{2-13}{11-13}
				\rectanglecolor{gray3}{13-13}{21-13}
				\Body
				\toprule
				&
				\multicolumn{3}{c}{\underline{\cart}} &
				\multicolumn{3}{c}{\underline{\acro}} &
				\multicolumn{3}{c}{\underline{\pendu}} &
				\multicolumn{3}{c}{\underline{\pusher}} \\

				&
				Max score $\uparrow$  & Rel. spread $\downarrow$ & Rel. iter. $\downarrow$ &
				Max score $\uparrow$  & Rel. spread $\downarrow$ & Rel. iter. $\downarrow$ &
				Max score $\uparrow$  & Rel. spread $\downarrow$ & Rel. iter. $\downarrow$ &
				Max score $\uparrow$  & Rel. spread $\downarrow$ & Rel. iter. $\downarrow$ \\
				
				\midrule
				
REI. & 499.3 & 0.24\%& 1485 &-154.0 & 81.88\%& 1499 &993.7 & 1.13\%& 1347 &-53.4 & 8.43\%& 2848 \\
$+ \lossU$ & 489.5\newline\textit{0.98x} & 1.84\%\newline\worse{\textit{7.65x}}& 1028\newline\better{\textit{0.69x}} &-98.1\newline\better{\textit{1.57x}} & 4.89\%\newline\better{\textit{0.06x}}& 803\newline\better{\textit{0.54x}} &975.8\newline\textit{0.98x} & 2.73\%\newline\worse{\textit{2.42x}}& 803\newline\better{\textit{0.60x}} &-49.6\newline\better{\textit{1.08x}} & 4.23\%\newline\better{\textit{0.50x}}& 763\newline\better{\textit{0.27x}} \\
$+ \lossG$ & 493.7\newline\textit{0.99x} & 0.95\%\newline\worse{\textit{3.96x}} & 1021\newline\better{\textit{0.69x}} &-98.5\newline\better{\textit{1.56x}} & 4.47\%\newline\better{\textit{0.05x}} & 479\newline\better{\textit{0.32x}} &979.6\newline\textit{0.99x} & 2.01\%\newline\worse{\textit{1.78x}} & 870\newline\better{\textit{0.65x}} &-50.2\newline\better{\textit{1.06x}} & 4.58\%\newline\better{\textit{0.54x}} & 1100\newline\better{\textit{0.39x}} \\

\midrule

REI. $+ V(s)$ & 492.6 & 1.64\%& 1299 &-190.3 & 80.03\%& 1499 &993.0 & 1.14\%& 1464 &-47.0 & 3.83\%& 2489 \\
$+ \lossU$ & 492.3\newline\better{\textit{1.00x}} & 1.73\%\newline\textit{1.05x}& 1302\newline\better{\textit{1.00x}} &-132.8\newline\better{\textit{1.43x}} & 77.41\%\newline\better{\textit{0.97x}}& 649\newline\better{\textit{0.43x}} &991.7\newline\better{\textit{1.00x}} & 1.47\%\newline\worse{\textit{1.29x}}& 1466\newline\better{\textit{1.00x}} &-46.6\newline\better{\textit{1.01x}} & 3.65\%\newline\better{\textit{0.95x}}& 2702\newline\textit{1.09x} \\
$+ \lossG$ & 490.6\newline\better{\textit{1.00x}} & 1.37\%\newline\better{\textit{0.83x}} & 1374\newline\textit{1.06x} &-114.5\newline\better{\textit{1.66x}} & 55.98\%\newline\better{\textit{0.70x}} & 723\newline\better{\textit{0.48x}} &990.0\newline\better{\textit{1.00x}} & 1.02\%\newline\better{\textit{0.90x}} & 1358\newline\better{\textit{0.93x}} &-46.9\newline\better{\textit{1.00x}} & 3.84\%\newline\better{\textit{1.00x}} & 2950\newline\worse{\textit{1.19x}} \\

\midrule

PPO & 474.5 & 8.13\%& 871 &-98.3 & 65.31\%& 1497 &961.2 & 4.09\%& 1356 &-40.0 & 1.25\%& 2999 \\
$+ \lossU$ & 474.9\newline\better{\textit{1.00x}} & 4.99\%\newline\better{\textit{0.61x}}& 778\newline\better{\textit{0.89x}} &-98.5\newline\better{\textit{1.00x}} & 65.99\%\newline\textit{1.01x}& 1479\newline\better{\textit{0.99x}} &963.8\newline\better{\textit{1.00x}} & 3.70\%\newline\better{\textit{0.91x}}& 1225\newline\better{\textit{0.90x}} &-40.8\newline\textit{0.98x} & 3.43\%\newline\worse{\textit{2.75x}}& 2966\newline\better{\textit{0.99x}} \\
$+ \lossG$ & 484.7\newline\better{\textit{1.02x}} & 4.39\%\newline\better{\textit{0.54x}} & 768\newline\better{\textit{0.88x}} &-77.0\newline\better{\textit{1.28x}} & 1.30\%\newline\better{\textit{0.02x}} & 494\newline\better{\textit{0.33x}} &969.7\newline\better{\textit{1.01x}} & 3.46\%\newline\better{\textit{0.85x}} & 929\newline\better{\textit{0.69x}} &-40.7\newline\textit{0.98x} & 2.95\%\newline\worse{\textit{2.36x}} & 2997\newline\better{\textit{1.00x}} \\

				\midrule
				
				&
				\multicolumn{3}{c}{\underline{\smash{\hopper}}} &
				\multicolumn{3}{c}{\underline{\walker}} &
				\multicolumn{3}{c}{\underline{\cheetah}} &
				\multicolumn{3}{c}{\underline{\ant}} \\

REI. & 331.5 & 52.64\%& 2235 &251.3 & 27.26\%& 1280 &993.5 & 60.25\%& 3934 &570.7 & 14.09\%& 688 \\
$+ \lossU$ & 423.2\newline\better{\textit{1.28x}} & 33.58\%\newline\better{\textit{0.64x}}& 3620\newline\worse{\textit{1.62x}} &364.1\newline\better{\textit{1.45x}} & 11.92\%\newline\better{\textit{0.44x}}& 798\newline\better{\textit{0.62x}} &1276.3\newline\better{\textit{1.28x}} & 29.37\%\newline\better{\textit{0.49x}}& 1731\newline\better{\textit{0.44x}} &754.4\newline\better{\textit{1.32x}} & 7.99\%\newline\better{\textit{0.57x}}& 71\newline\better{\textit{0.10x}} \\
$+ \lossG$ & 474.8\newline\better{\textit{1.43x}} & 50.88\%\newline\better{\textit{0.97x}} & 1877\newline\better{\textit{0.84x}} &356.2\newline\better{\textit{1.42x}} & 12.13\%\newline\better{\textit{0.44x}} & 702\newline\better{\textit{0.55x}} &1251.8\newline\better{\textit{1.26x}} & 38.11\%\newline\better{\textit{0.63x}} & 1806\newline\better{\textit{0.46x}} &776.6\newline\better{\textit{1.36x}} & 11.22\%\newline\better{\textit{0.80x}} & 96\newline\better{\textit{0.14x}} \\

\midrule

REI. $+ V(s)$ & 397.1 & 16.39\%& 4711 &295.5 & 26.77\%& 1270 &1609.8 & 40.37\%& 4589 &524.4 & 59.82\%& 4604 \\
$+ \lossU$ & 642.5\newline\better{\textit{1.62x}} & 35.42\%\newline\worse{\textit{2.16x}}& 2430\newline\better{\textit{0.52x}} &416.0\newline\better{\textit{1.41x}} & 13.99\%\newline\better{\textit{0.52x}}& 1056\newline\better{\textit{0.83x}} &1542.0\newline\textit{0.96x} & 31.98\%\newline\better{\textit{0.79x}}& 4987\newline\textit{1.09x} &828.7\newline\better{\textit{1.58x}} & 9.87\%\newline\better{\textit{0.17x}}& 34\newline\better{\textit{0.01x}} \\
$+ \lossG$ & 734.6\newline\better{\textit{1.85x}} & 16.58\%\newline\textit{1.01x} & 1835\newline\better{\textit{0.39x}} &461.5\newline\better{\textit{1.56x}} & 22.19\%\newline\better{\textit{0.83x}} & 1447\newline\worse{\textit{1.14x}} &1593.2\newline\textit{0.99x} & 24.70\%\newline\better{\textit{0.61x}} & 4352\newline\better{\textit{0.95x}} &825.3\newline\better{\textit{1.57x}} & 10.15\%\newline\better{\textit{0.17x}} & 5\newline\better{\textit{0.00x}} \\

\midrule

PPO & 814.0 & 20.57\%& 2794 &518.4 & 24.42\%& 2508 &2021.4 & 30.50\%& 2999 &775.2 & 9.16\%& 2874 \\
$+ \lossU$ & 980.0\newline\better{\textit{1.20x}} & 24.84\%\newline\worse{\textit{1.21x}}& 1780\newline\better{\textit{0.64x}} &741.0\newline\better{\textit{1.43x}} & 34.51\%\newline\worse{\textit{1.41x}}& 1670\newline\better{\textit{0.67x}} &1851.0\newline\textit{0.92x} & 26.65\%\newline\better{\textit{0.87x}}& 2999\newline\better{\textit{1.00x}} &857.3\newline\better{\textit{1.11x}} & 7.27\%\newline\better{\textit{0.79x}}& 1337\newline\better{\textit{0.47x}} \\
$+ \lossG$ & 838.6\newline\better{\textit{1.03x}} & 32.73\%\newline\worse{\textit{1.59x}} & 2361\newline\better{\textit{0.85x}} &676.1\newline\better{\textit{1.30x}} & 26.77\%\newline\textit{1.10x} & 1706\newline\better{\textit{0.68x}} &1998.1\newline\textit{0.99x} & 20.79\%\newline\better{\textit{0.68x}} & 2999\newline\better{\textit{1.00x}} &844.2\newline\better{\textit{1.09x}} & 6.11\%\newline\better{\textit{0.67x}} & 1366\newline\better{\textit{0.48x}} \\

				\bottomrule
			\end{NiceTabular}
			\vspace{-1pt}
		\end{table*}

\subsection{Results}
\label{sec:results}
All settings, each algorithm variation for each environment, were run $20$ times.
The results are summarized in Figures~\ref{fig:profiles} and Table~\ref{table:results}.
In Figure~\ref{fig:profiles} we show the performance profiles~\cite{agarwalDeepReinforcementLearning2021} of the different algorithms, which is a recommended summary measure to compare RL methods on multiple environments/tasks.
The plot shows the fraction of runs that achieve a certain relative normalized score using all runs in all environments, which allows a qualitative comparison and shows all score percentiles.
The relative normalized score is calculated for each environment by using min-max normalization, with min and max computed over all algorithms and the min/max taken over the max scores averaged over $100$ iterations.
As these performance profiles show a summarized evaluation over all runs in all environments, we can see that the (1) the proposed backpropagation-free method can compete with the backpropagation-based baselines in terms of performance, and (2) that the proposed methods improve stability and consistency during training as their score distributions are higher especially in the lower score regimes,
i.e., fewer runs got stuck in bad local optima.
In the high performance regions ($\tau > 0.98$) we see that the backpropagation-based REINFORCE achieves a slightly higher peak performance.
Indicating having few higher performing runs at the cost of more bad performing runs and stability.

In Figure~\ref{fig:eval} we show the performance for each environment and algorithm.
Normalized scores are calculated by averaging the smoothed raw scores (cumulative reward) and using the $5\%$ and $95\%$ percentiles as min and max for min-max normalization.
This normalization allows for relative comparisons of the methods in the environments.
Across the different comparisons and environments, we see that the proposed method can achieve at least comparable performance with their BP-counterparts in all settings (as summarized before with the performance profiles).
In addition, especially in the more complex environments (\hopper, \walker, \cheetah, and \ant) where there is no defined maximal performance, we often see an increased performance.
The improvement in stability and consistency is also reflected in the lower spread.
In some environments, it does come at the cost of speed in terms of required iterations.
However, this gap is most prominent in the easier environments (\cart, \acro, and \pendu) that do not require many iterations in general, whereas in the remaining environments that gap is not present.

The aforementioned improvement in stability and consistency as well as the speed differences are evaluated in more detail in Table~\ref{table:results}.
Here we show the (1) \inlineBoxPad{gray1}{\textit{Max score}}: the maximal mean score averaged over $100$ iterations, (2) \inlineBoxPad{gray2}{\textit{Rel. spread}}: the relative spread in $\%$ around this max score as the $95\%$ CI divided by \textit{Max score}, and (3) \inlineBoxPad{gray3}{\textit{Rel. iter.}}: the relative iteration in which the max score was achieved in relation to the baseline.
Additionally we show and highlight the relative changes in the these metrics for easier comparison and colored them in \better{\textit{green}} if the proposed method is at least as good as the baseline, and \worse{\textit{red}} if it is more than \textit{1.1x} worse.

Supporting the previous discussion, the \inlineBoxPad{gray1}{\textit{Max score}} is equal or higher than the baseline in $38/48$ cases, with $8$ cases only slightly worse $>=$ \textit{0.98x}, and two \textit{outliers} with \textit{0.96x} and \textit{0.92x}.
Comparing the  \inlineBoxPad{gray2}{\textit{Relative spread}} we see that in $33/48$ cases the spread is smaller.
Where the \textit{relative spread} is higher, we either see low absolute spread (e.g., \cart~or \pendu) or a combination with stronger performance in unbounded environments (e.g., \hopper~or \walker).
The comparisons of \inlineBoxPad{gray3}{\textit{Relative iterations}}, i.e., the learning speed, supports the statement from above, that the learning speed gap is mostly not existent in the more challenging environments and relative small in the easier ones.

\begin{tcolorbox}[size=small, colback=gray3,colframe=gray3, boxsep=1pt]
In summary, the presented evaluations show that the proposed backpropagation-free method can compete with the backpropagation-based baselines, while often improving performance and stability -- especially improving consistency of the algorithms, having fewer runs stuck in bad local optima.
\end{tcolorbox}

\section{Limitations \& Discussion}
\label{sec:discussion}
The evaluations demonstrated that our proposed method offers a promising approach to backpropagation-free training of neural policies, compatible with various reinforcement learning algorithms and improving their learning behavior. 
However, there are open questions, potential limitations regarding the scalability and transferability of this approach, and the assumption of a global target.

\paragraph{Scalability:} One potential limitation is scaling in terms of network depth and batch size $N$. 
The pairwise distance matrices, which are $N \times N$, may become computationally expensive as $N$ increases, especially with full episode based updates. 
A potential solution to this issue is the use of sparse matrices, only considering a certain distance neighborhood, to reduce computational complexity and focus on small neighborhoods.
See Appendix~\ref{sec:sparse_matrices} for initial results applying this concept.
With increasing network depth, i.e., more layers trained with the pairwise distance based loss, it needs to be investigated if useful representations are still learned.

\paragraph{Global target:} For calculating the local losses, we used $\bar{D}_{X}$ as the target distance matrix for all layers, assuming a global target available to all layers. 
This assumption may be circumvented by, for example, exploring random transformations as targets~\cite{noklandTrainingNeuralNetworks2019} or additional forward-error propagation~\cite{kohanErrorForwardPropagationReusing2018}.
See Appendix~\ref{sec:forward_error} for initial results applying the idea of forward-error propagation.

\paragraph{Frozen Random Layers}
To evaluate if the observed benefit is due to learning meaningful representations in higher dimensions, and not due to the mapping to a higher dimension itself, we also trained agents using frozen random hidden layers. 
The results (see Appendix~\ref{sec:frozen_hidden}) indicate that the proposed approach is able to learn meaningful transformations into higher dimensions.

\paragraph{Transferability:} Another potential limitation is the transferability of the approach to different network architectures, such as convolutional layers. 
The choice of the underlying distance metric plays a crucial role in this transferability. 
Here, we used the $\ell_1$-norm as a distance measurement suitable for higher dimensions~\cite{aggarwalSurprisingBehaviorDistance2001} and which worked best in preliminary tests.
Calculating all pairwise distances can become expensive with larger batch sizes (e.g., the proposed approach was on average $1.5x$ slower per iteration) and investigating metrics that are suitable for fast computation via matrix operations like $\ell_2$-norm or gram matrix based metrics is a useful future direction.
Additionally, for other model architectures, specialized distance or similarity measurements might be more beneficial. 
Task-specific adaptations can also be incorporated, similar to our proposed guided loss $\lossG$, which scales the pairwise distances with rewards or performance-related feedback. 
The flexibility of our method allows for the use of various distance measurements tailored to specific tasks or domains.

\paragraph{Hyperparameters:} As with all deep (reinforcement) learning methods, hyperparameters are critical. 
Besides the general parameters influencing the underlying RL algorithm (Section~\ref{sec:setup_parameters}), the scaling of the distance matrices can also be considered hyperparameters. 
The choice of normalization and scaling should be adapted to different environments or tasks, much like the selection of distance measurements.
The here used scaling might not be the best fit in all scenarios. 
Our approach's generality allows for such adaptations, enhancing its applicability across various settings.

\paragraph{Potential Benefits and Future Directions:} Beyond the demonstrated benefits, the layer-wise unsupervised loss introduces a promising avenue for transferring or sharing learned representations. 
This could be particularly useful in multi-agent setups~\cite{silvaSurveyTransferLearning2019,maRecursiveReasoningGraph2022,rotherDisentanglingInteractionUsing2023}, meta-RL~\cite{beckSurveyMetaReinforcementLearning2023}, or continual learning~\cite{verwimpContinualLearningApplications2023}, where learned hidden layers can be shared while training different output layers accordingly. 
In settings where an offline dataset is available, the proposed method may be used for offline pre-training of the representation layers~\cite{yang2021representation}. 
This might help the RL agent to learn more efficiently and robustly.
Additionally, layer-wise training allows for different learning rates for learning representations (hidden layers) and behavior (output layer), accommodating various learning speeds~\cite{marionLeveragingTwotimescaleRegime2023}.
Initial results on mixing different learned speeds are presented in Appendix~\ref{sec:diff_learnig_speeds}.
Since our approach does not require backpropagation, the entire network does not need to be fully differentiable or even fully known. 
This opens up interesting possibilities for incorporating black-box operations between layers~\cite{hintonForwardForwardAlgorithmPreliminary2022a}, further expanding the method's versatility.
\begin{tcolorbox}[size=small, colback=gray3,colframe=gray3]
In summary, while our backpropagation-free method shows promising results and a flexible framework, future research should challenge its scalability and transferability, explore different metrics and scalings for different environments, and investigate its application within other network architectures and reinforcement learning algorithms, e.g., value-based methods.
\end{tcolorbox}

\section{Conclusion}
\label{sec:conclusion} 
In this paper, we introduced a novel approach for backpropagation-free training of neural networks in reinforcement learning settings. 
Our method focuses on learning feature transformations by minimizing the difference between pairwise distance matrices in the hidden layers, leveraging the concept of multi-dimensional scaling.
We demonstrated the effectiveness of our approach across a set of common RL benchmarks, showing that it is compatible with various RL algorithms. 
Our method not only matched but often improved the performance of traditional backpropagation-based training, while also enhancing training stability and consistency.

The versatility of the proposed method invites future research into utilizing different distance metrics and non-linear dimensionality reduction techniques. 
Additionally, it offers potential for leveraging the task-independent representation learning for transfer.
As the interest in backpropagation-free alternatives for training neural networks regains attraction~\cite{ororbiaBackpropFreeReinforcementLearning2022,hintonForwardForwardAlgorithmPreliminary2022a,ororbiaPredictiveForwardForwardAlgorithm2023,guan2024temporal}, we hope this work inspires new ideas and research in alternative learning approaches, particularly within the underexplored domain of reinforcement learning.





\bibliographystyle{unsrtnat}
\bibliography{FFRL}

\newpage
\onecolumn
\appendix
\renewcommand{\thefigure}{A\arabic{figure}}
\setcounter{figure}{0}
\setcounter{section}{0}
\section*{Appendix}
All plots show the mean (solid line) and $95\%$ CI (shaded area) of the normalized score over $20$ runs each and smoothed over $100$ iterations.

\section{Scalability -- Sparse Distance Matrices}
\label{sec:sparse_matrices}
\begin{figure}[H]
	\includegraphics[width=\linewidth]{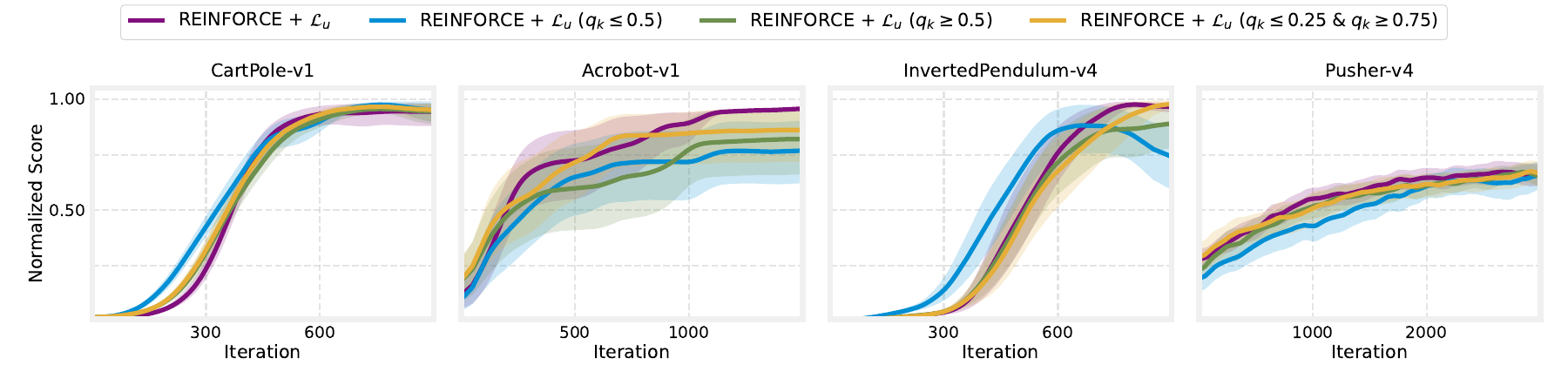}
	\caption{
		Comparing the backpropagation-free method $+ \lossU$ using the full pairwise distance matrix against utilizing sparse distance matrices with only half of the entries per data point.
	}
	\label{fig:neighborhood}
\end{figure}

To counter the increased computational complexity due to increased batch size $N$, one possible solution is to use sparse pairwise distance matrices by only looking at certain neighborhoods of each data point.
Figure~\ref{fig:neighborhood} shows initial results applying this idea.
Here, we only kept half of the pairwise distances for each data point by discarding distances based on different percentiles.
With (1) $q_k \leq 0.5$ we only kept the closest half of neighbors, with (2) $q_k \geq 0.5$ we kept the furthest half, and with (3) $q_k \leq 0.25$ \& $q_k \geq 0.75$ we kept the closest and furthest quarter.
The combination of considering the closest and furthest neighbors (3) seems to be a promising candidate for a trade-off between learning speed, performance, and computational complexity.

\section{Global Target -- Forward-error Propagation}
\label{sec:forward_error}
\begin{figure}[H]
	\includegraphics[width=\linewidth]{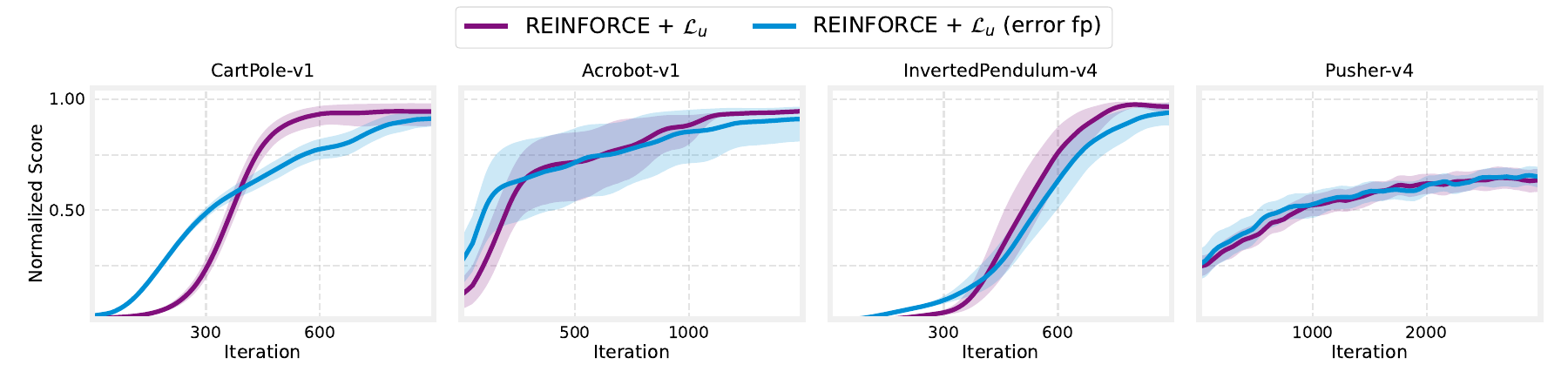}
	\caption{
		Comparing the backpropagation-free method $+ \lossU$ using the global target against the forward-error propagation idea (error fp). 
	}
	\label{fig:forward_error}
\end{figure}

In the proposed method, the target pairwise distance matrix is known to all layers as a global target.
One possibility to remove this assumption is to use an idea similar to forward-error propagation~\cite{kohanErrorForwardPropagationReusing2018}.
For that, each hidden layer propagates its error $e_l$ -- here, the differences in the pairwise distances -- to the next layer in addition to the \textit{normal} feedforward information $h_l$.
The target distance matrix $D_X$ is then calculated from the input of the previous layer (instead of the input data) and we use the previous error to scale this matrix (before its described normalization).
For that, the error is first normalized to be within $[1,2]$ with $e_l = \frac{\abs{e_l}}{\max\;\abs{e_l}} + 1$, and then used to scale the target distance matrix $D_X = D_X \cdot e_l$.
Results using this error forward-propagation (error fp) are shown in Figure~\ref{fig:forward_error}.
The method is able to learn successful policies.
Additionally, these initial results suggest that concentrating on the errors of previous layers may increase initial learning speed and performance.

\newpage
\section{Different Learning Speeds}
\label{sec:diff_learnig_speeds}
\begin{figure}[H]
	\includegraphics[width=\linewidth]{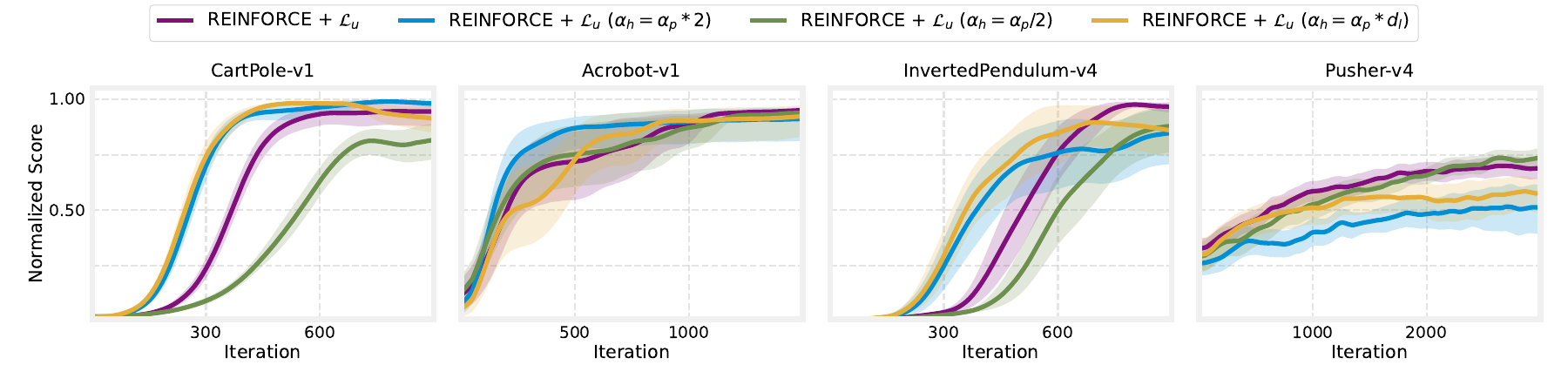}
	\caption{
		Comparing the backpropagation-free method $+ \lossU$ with the same learning rate for all layers against variations of mixed learning rates.
	}
	\label{fig:diff_learning_speeds}
\end{figure}

As all layers are trained independently with local losses -- technically with separate instantiates of the optimizer  -- it is easy to use different learning rates $\alpha$ for the different layers.
We denote the learning rates of the hidden layers and the output (policy) layer by $\alpha_h$ and $\alpha_p$ respectively.
In Figure~\ref{fig:diff_learning_speeds} we show three simple approaches to utilize different learning speeds for the different layers, i.e., for the feature mapping learning and policy learning: 
(1) higher learning speed for the hidden layers ($\alpha_h = \alpha_p \cdot 2$), 
(2) lower learning speed for the hidden layers ($\alpha_h = \alpha_p / 2$), and 
(3) higher learning speed for the hidden layers scaled by their \textit{distance} to the policy layer ($\alpha_h = \alpha_p \cdot d_l$), where $d_l = 2$ for the layer right before the policy layer, $d_l = 2$ for the one before, and so on.
The initial results show that different combinations of learning speeds can be beneficial depending on the environment.

\section{Frozen Random Layers}
\label{sec:frozen_hidden}
\begin{figure}[H]
	\includegraphics[width=\linewidth]{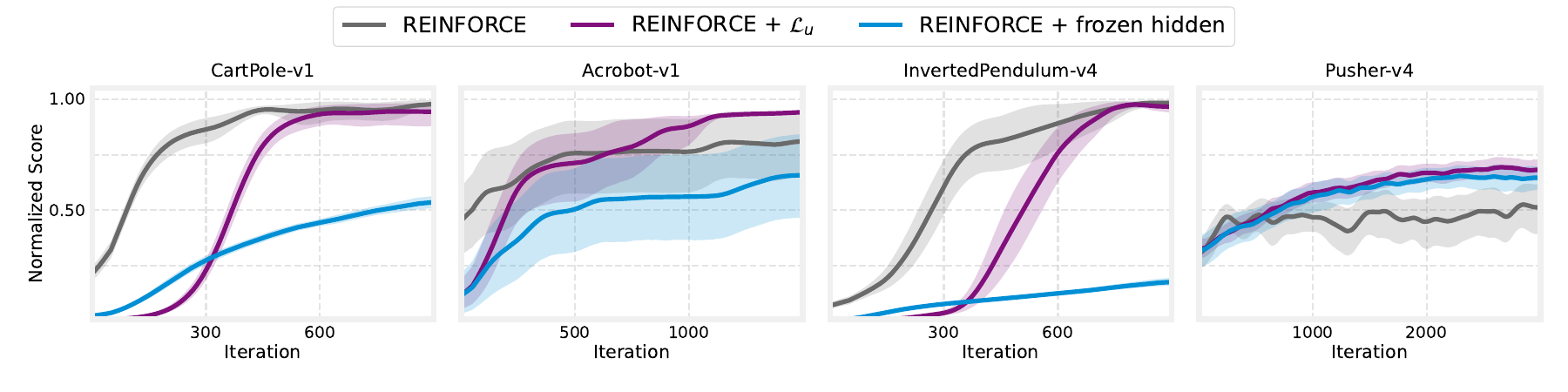}
	\caption{
		Comparing the baseline, backpropagation-free method $+ \lossU$, and frozen random hidden layers.
	}
	\label{fig:frozen}
\end{figure}

To evaluate that the proposed method learns a \textit{helpful / meaningful} transformation into higher dimensions, we compare it against using frozen random hidden layers.
The experiments summarized in Figure~\ref{fig:frozen} indicate that the learned structure by the proposed method, rather than mere high-dimensional transformation, is responsible for the reported improvements.
The strength of the effect depends on the environment. 

\end{document}